%% file: main.tex
\documentclass[nohyperref]{article}

\usepackage{microtype}
\usepackage{graphicx}
\usepackage{subfigure}
\usepackage{booktabs} % for professional tables
\usepackage{amssymb}
\usepackage{etoolbox,refcount}
\usepackage{multicol}
\usepackage{xcolor}
\usepackage{bm}

\usepackage{hyperref}

\usepackage[accepted]{icml2023}

\usepackage{amsmath}
\usepackage{amssymb}
\usepackage{mathtools}
\usepackage{amsthm}

\usepackage{multirow}
\usepackage{enumitem}
\usepackage[capitalize,noabbrev]{cleveref}

\theoremstyle{plain}

\theoremstyle{definition}

\theoremstyle{remark}

\usepackage[textsize=tiny]{todonotes}

\icmltitlerunning{Heterogeneous Federated Learning Using Knowledge Codistillation}

\begin{document}

\twocolumn[
\icmltitle{Heterogeneous Federated Learning Using Knowledge Codistillation}



\icmlsetsymbol{equal}{*}

\begin{icmlauthorlist}
\icmlauthor{Jared Licharge}{comp}
\icmlauthor{Ehsan Amid}{brain}
\icmlauthor{Shankar Kumar}{comp}
\icmlauthor{Tien-Ju Yang}{comp}
\icmlauthor{Rohan Anil}{brain}
\icmlauthor{Rajiv Mathews}{comp}

\end{icmlauthorlist}

\icmlaffiliation{comp}{Google Research}
\icmlaffiliation{brain}{Google Brain}

\icmlcorrespondingauthor{Jared Lichtarge}{lichtarge@google.com}
\icmlcorrespondingauthor{Ehsan Amid}{eamid@google.com}

\icmlkeywords{Machine Learning, ICML}

\vskip 0.3in
]



\input{defs}
\printAffiliationsAndNotice{}  % leave blank if no need to mention equal contribution

\begin{abstract}
Federated Averaging, and many federated learning algorithm variants which build upon it, have a limitation: all clients must share the same model architecture. This results in unused modeling capacity on many clients, which limits model performance. To address this issue, we propose a method that involves training a small model on the entire pool and a larger model on a subset of clients with higher capacity. The models exchange information bidirectionally via knowledge distillation, utilizing an unlabeled dataset on a server without sharing parameters. We present two variants of our method, which improve upon
federated averaging on image classification and language modeling tasks. We show this technique can be useful even if only out-of-domain or limited in-domain distillation data is available. Additionally, the bi-directional knowledge distillation allows for domain transfer between the models when different pool populations introduce domain shift.

% Federated Averaging, the most widely used algorithm for Federated Learning, has a limitation: all clients must share the same model architecture, which means that the model size is restricted to the smallest size that can fit on all clients in the federated pool. This results in unused modeling capacity on many clients, negatively impacting model performance. To address this issue, we propose a method that involves training a small model on the entire pool and a larger model on a subset of clients with higher capacity. Information is shared between the models through bi-directional knowledge distillation using an unlabeled server dataset without sharing parameters. Two variants of our method, \inter\ and \merged, improve upon the standard \fedavg\ model on image classification and language modeling tasks. Additionally, the bi-directional knowledge distillation allows for domain transfer between the models when different pool populations introduce domain shift.
\end{abstract}

\input{1_introduction}
\input{2_method}
\input{5_experiments}

\input{6_discussion}
\input{7_conclusion}

% % Acknowledgements should only appear in the accepted version.
% \section*{Acknowledgements}

% \textbf{Do not} include acknowledgements in the initial version of
% the paper submitted for blind review.

% If a paper is accepted, the final camera-ready version can (and
% probably should) include acknowledgements. In this case, please
% place such acknowledgements in an unnumbered section at the
% end of the paper. Typically, this will include thanks to reviewers
% who gave useful comments, to colleagues who contributed to the ideas,
% and to funding agencies and corporate sponsors that provided financial


\bibliography{example_paper}
\bibliographystyle{icml2023}

%%%%%%%%%%%%%%%%%%%%%%%%%%%%%%%%%%%%%%%%%%%%%%%%%%%%%%%%%%%%%%%%%%%%%%%%%%%%%%%
%%%%%%%%%%%%%%%%%%%%%%%%%%%%%%%%%%%%%%%%%%%%%%%%%%%%%%%%%%%%%%%%%%%%%%%%%%%%%%%
% APPENDIX
%%%%%%%%%%%%%%%%%%%%%%%%%%%%%%%%%%%%%%%%%%%%%%%%%%%%%%%%%%%%%%%%%%%%%%%%%%%%%%%
%%%%%%%%%%%%%%%%%%%%%%%%%%%%%%%%%%%%%%%%%%%%%%%%%%%%%%%%%%%%%%%%%%%%%%%%%%%%%%%
\newpage
\appendix
\onecolumn

\input{8a_related_work}
\input{8b_app_methods}

\input{8c_experiment_setup}
\input{8d_experiments}

\input{8z_appendix}
%%%%%%%%%%%%%%%%%%%%%%%%%%%%%%%%%%%%%%%%%%%%%%%%%%%%%%%%%%%%%%%%%%%%%%%%%%%%%%%
%%%%%%%%%%%%%%%%%%%%%%%%%%%%%%%%%%%%%%%%%%%%%%%%%%%%%%%%%%%%%%%%%%%%%%%%%%%%%%%

\end{document}

%% file: defs.tex
\let\oldemptyset\emptyset

\newcommand{\inter}{{\sc PeriodicCodist}}
\newcommand{\merged}{{\sc MergedCodist}}
\newcommand{\fedavg}{{\sc FedAvg}}

\newcommand{\pool}{\mathcal{P}}
\newcommand{\mainmodel}{\theta_{H}}
\newcommand{\auxmodel}{\theta_{L}}
\newcommand{\umainmodel}{\theta_H}
\newcommand{\uauxmodel}{\theta_{L}}
\newcommand{\mainpool}{\pool_{H}}
\newcommand{\auxpool}{\pool_{L}}

\newcounter{countitems}
\newcounter{nextitemizecount}
\newcommand{\setupcountitems}{%
  \stepcounter{nextitemizecount}%
  \setcounter{countitems}{0}%
  \preto\item{\stepcounter{countitems}}%
}
\makeatletter
\newcommand{\computecountitems}{%
  \edef\@currentlabel{\number\c@countitems}%
  \label{countitems@\number\numexpr\value{nextitemizecount}-1\relax}%
}
\newcommand{\nextitemizecount}{%
  \getrefnumber{countitems@\number\c@nextitemizecount}%
}
\newcommand{\previtemizecount}{%
  \getrefnumber{countitems@\number\numexpr\value{nextitemizecount}-1\relax}%
}
\makeatother    
\newenvironment{AutoMultiColItemize}{%
\ifnumcomp{\nextitemizecount}{>}{3}{\begin{multicols}{2}}{}%
\setupcountitems\begin{itemize}}%
{\end{itemize}%
\unskip\computecountitems\ifnumcomp{\previtemizecount}{>}{3}{\end{multicols}}{}}

\definecolor{pythonblue}{RGB}{0, 0, 255}
\definecolor{pythongreen}{RGB}{0, 127.5, 0}

%% file: 1_introduction.tex
\section{Introduction}\label{sec:intro}

Federated Learning (FL) \citep{mcmahan2016federated} is a technique for training a centralized model using the private data of decentralized clients without the need for data sharing. It is typically implemented through the Federated Averaging Algorithm \cite{mcmahan2017communication} (\fedavg), which combines model updates from a subset of clients in each round to produce a single update on the shared server model. However, this approach requires all clients to use the same model architecture. As the client pool is often composed of diverse hardware with varying capacities, sharing an architecture limits the shared model size to the lowest-common-denominator architecture of the federated pool. This results in unused modeling capacity on many clients, which detriments model performance. To overcome this limitation, newer federated learning methods are being developed \citep{lin2020, ozkara2021quped} that remove the constraint that all models in the pool must share a single architecture. While some methods alleviate this constraint by soliciting logits from the clients instead of model differences \cite{lin2020}, we seek to address this issue by building on top of the popular \fedavg\ strategy. A thorough discussion of related work is presented in Appendix \ref{app:related_work}.
%The simplest strategy of doing so would be to partition the set of clients into two (or more) pools, sorted by model-size capacity, and run \fedavg\ independently on each pool using the lowest-common-denominator model of that pool. However, this silos information between the groups, which hurts model performance and introduces domain shift (e.g., the larger model sees more data from devices with higher capacity). We propose to overcome this information silo problem by 
In this work, we propose to partition the set of clients into two or more pools by model-size capacity, and run (bi-directional) knowledge distillation \cite{hinton2015distilling}, known as \emph{codistillation}~\citep{anil2018}, between the server models produced by each \fedavg\ pool, leveraging (un)labeled server data as the distillation dataset. By co-distilling the two (or more) models frequently over the course of \fedavg\ rounds, we can share information between the pools without sharing model parameters. This partitioning allows us to make fuller use of the computational capacity of the federated pool, which can lead to increased performance and convergence in fewer federated rounds. Our methods are agnostic to the FL method applied within each pool and can work in a complementary fashion with any of the variants of \fedavg\ which still train a single server model\footnote{See related work discussion in Appendix \ref{app:related_work}.}. Our approach places no additional computational burden on the client, as all distillation computation occurs on the server.

%% file: 2_method.tex
\begin{figure}[]
\begin{center}
% \vspace{-0.2cm}
    \subfigure[]{\includegraphics[width=0.49\linewidth]{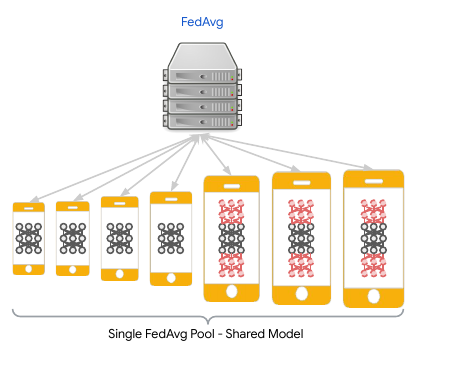}}
    % \hspace{-0.5cm}
    \subfigure[]{\includegraphics[width=0.49\linewidth]{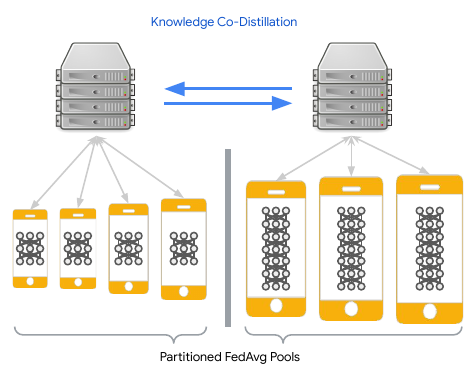}}
    % \hspace{0.3cm}
    % \subfigure[]{\includegraphics[width=0.24\linewidth]{figs/periodic.pdf}}
    % % \hspace{0.1cm}
    % \subfigure[]{\includegraphics[width=0.24\linewidth]{figs/merged.pdf}}
    % \vspace{-0.3cm}
    \caption{Schematics of the proposed methods: (a) Baseline \fedavg\ on the single pool of clients. The model capacity is limited by the lowest-common-denominator client's capacity in the pool. (b) Federated co-distillation with two client pools. The larger model trains on a smaller pool of high-capacity clients. The models exchange information via co-distillation on the server. %(c) An illustration of the \inter\ approach in the weight space. After $p=3$ \fedavg\ steps, we apply a round of co-distillation for $s=15$ steps by using the other model as a teacher. The \fedavg\ resumes after the co-distillation round. (d) An illustration of the \merged\ approach. In each round, we merge the scaled distillation direction (by calculating the distillation direction after $s=15$ steps) with the \fedavg\ direction before applying the update to the model.}
    }
    \label{fig:methods}
    \vspace{-0.3cm}
    \end{center}
\end{figure}

\section{Methods}\label{sec:method}

% Federated Learning (FL) is most commonly implemented through Federated Averaging (\fedavg)~\citep{mcmahan2016federated} or similar methods~\citep{wang2021field}. \fedavg\ works by combining model updates from clients within the same federated pool but this requires using the same server model for all clients regardless of their hardware capacity. To make better use of larger hardware, clients can be grouped into separate \fedavg\ pools based on capacity but this leads to siloing information from clients with lower capacity.
We present two novel federated learning algorithms that avoid the requirement for a common server model by using knowledge distillation to share information between the server models of different \fedavg\ pools without sharing model parameters. We present two variants of our method: \inter\, in which co-distillation is applied independently from \fedavg\ updates at fixed intervals of federated rounds, and \merged, in which co-distillation updates are merged with \fedavg\ updates at each round. Neither approach is restricted to a specific distillation technique, as any underlying knowledge distillation strategy can be employed~\citep{hinton2015distilling,romero2014fitnets,tian2019contrastive,muller2020subclass,amid2022layerwise}. In this work, we use KL-divergence between output probabilities for performing distillation. We explore using two federated pools, one for low-capacity and another for high-capacity models, but our approach can be extended to work for any number of pools. The standard \fedavg\ method and our proposed co-distillation approaches are illustrated in Figure~\ref{fig:methods}(a) and Figure~\ref{fig:methods}(b), respectively. 

We denote by $\pool$ the pool of clients with heterogeneous memory constraints. For ease of presentation, we consider the case where $\pool$ is partitioned into two pools: the high-performance clients $\mainpool$ having devices with higher capacity, and the remaining (low-performance) clients $\auxpool = \pool - \mainpool$. Based on the distribution of the devices across a typical federated client pool~\cite{wang2021field}, it is reasonable to assume that $\vert \mainpool\vert \ll \vert \auxpool\vert$. We consider training two model architectures, parameterized by $\auxmodel$ and $\mainmodel$, using the available clients in $\pool$, with $\vert\auxmodel\vert < \vert\mainmodel\vert$. Therefore, we assume that 
$\mainmodel$ is only deployable on the high-performance clients in $\mainpool \subset \pool$, while $\auxmodel$ can be trained on any client in $\pool$. In particular, we assume that $\auxmodel$  represents the lowest-common-denominator model size given the hardware constraints on the entire pool $\pool$. Therefore, vanilla \fedavg\ may be applied to train $\auxmodel$ via \fedavg\ on the entire pool $\pool$. However, $\mainmodel$ is not trainable on the client devices in $\auxpool$ due to resource constraints. %We also explore a scenario where we train $\auxmodel$ only on $\auxpool \subset \pool$.
$\mainmodel$ offers an advantage over $\auxmodel$ by having a higher capacity, while $\auxmodel$ benefits from a significantly larger pool of clients for training. Thus, model capacity imposes a trade-off in terms of modeling power and data availability. We aim to tackle this issue by allowing the larger model to benefit from the low-performance clients' data while allowing the smaller model to utilize the improved modeling power offered by the larger model as a teacher.

\subsection{Periodic Codistillation ({\bf \inter})}
In our first approach, we apply vanilla \fedavg\ for training each model on its corresponding available client pool. 
%The improvement offered by our approach over vanilla \fedavg\ relies on exchanging information between $\auxmodel$ and $\mainmodel$ via periodic codistillation rounds in between \fedavg\ rounds.
%Specifically, 
Let $T$ denote the total number of \fedavg\ rounds for training $\auxmodel$ and $\mainmodel$. In our \inter\ approach, we perform vanilla \fedavg\ on $\auxmodel$ and $\mainmodel$ using the pool of clients $\pool$ and $\mainpool \subset \pool$, respectively. However, we pause the process every $p$ rounds and perform codistillation for $s$ steps. The codistillation round involves creating a copy of each model $\auxmodel^t$ and $\mainmodel^t$ at round $t = kp$, where $k \in \mathbb{N}$, to serve as a teacher for distilling information to the other model. %Our \inter\ approach is not restricted by the distillation technique, and any underlying knowledge distillation strategy can be employed~\citep{hinton2015distilling,romero2014fitnets,tian2019contrastive,muller2020subclass,amid2022layerwise}.

Let $\uauxmodel^{\text{\tiny student}}$ and $\umainmodel^{\text{\tiny student}}$ respectively denote the updated student models after codistillation, starting from $\auxmodel^t$ and $\mainmodel^t$ at round $t$, using teachers $\umainmodel^{\text{\tiny teacher}}$ and $\uauxmodel^{\text{\tiny teacher}}$. At the end of the codistillation round, we replace $\auxmodel^t$ ($\mainmodel^t$) by $\uauxmodel^{\text{\tiny student}}$ ($\umainmodel^{\text{\tiny student}}$) and continue training the models using \fedavg\ at round $t+1$. The \inter\ strategy is summarized in Algorithm~\ref{alg:periodic} and illustrated in Figure \ref{fig:methods_illustrations}(a) in Appendix \ref{app:methods}.

% MOVED TO APPENDIX
% \begin{algorithm}[tb]
%   \caption{\inter: Periodic Codistillation}
%   \label{alg:periodic}
% \begin{algorithmic}[1]
%   \INPUT number of \fedavg\ rounds $T$, total clients pool $\pool$ and the subset of high-performance clients $\mainpool \subset \pool$, codistillation period $p$, codistillation steps $s$
%   \\\vspace{-0.25cm}\hrulefill
%   \vspace{-0.05cm}
%   \STATE Initialize parameters $\auxmodel^0$ and $\mainmodel^t$
%   \FOR{$t=1$ {\bfseries to} $T$}
%   \STATE Apply \fedavg\ on $\auxmodel^t$ and $\mainmodel^t$ by sampling clients from $\pool$ and $\mainpool$, respectively
%   \IF{$t\, \%\, p == 0$}
%   \STATE Create \textbf{teachers} $\auxmodel^{\text{\tiny teacher}} \gets \auxmodel^t$ and $\mainmodel^{\text{\tiny teacher}} \gets \mainmodel^t$
%   \STATE Initialize \textbf{students} $\uauxmodel^{\text{\tiny student}} \gets \auxmodel^t, \,\umainmodel^{\text{\tiny student}} \gets \mainmodel^t$
%   \STATE \textbf{In parallel}, distill to $\theta^{\text{\tiny student}}_{i}$ for $s$ steps using $\theta^{\text{\tiny teacher}}_{\{L, H\} - \{i\}}$ as the teacher, where $i \in \{L, H\}$
%   \STATE Replace $\auxmodel^t \gets \uauxmodel^{\text{\tiny student}}, \,\mainmodel^t \gets \umainmodel^{\text{\tiny student}}$
%   \ENDIF
%   \ENDFOR
%   \OUTPUT $\auxmodel^T$, $\mainmodel^T$
% \end{algorithmic}
% \vspace{-0.1cm}
% \end{algorithm}

% MOVED TO APPENDIX

\subsection{Merged Gradient Codistillation ({\bf \merged})}
We consider an alternative strategy where we apply codistillation in conjunction with \fedavg\ at every round. To motivate our strategy, we note that \fedavg\ aggregates \emph{model differences} from the clients participating in the round and treats the combined model difference as a single gradient to apply the update. We use a similar approach to treat the model difference obtained from applying a round of distillation as a gradient. By merging the two gradients, we can apply a single update to the model parameters at each round.

Similar to \inter, we conduct \fedavg\ independently for $\auxmodel$ and $\mainmodel$ using their respective pools of clients $\pool$ and $\mainpool \subset \pool$. We denote the gradients obtained for the models $\auxmodel$ and $\mainmodel$ by aggregating the client gradients at round $t$, as in vanilla \fedavg, as $g^t_L$ and $g^t_H$, respectively.
Before applying the \fedavg\ step at each round using the aggregated client gradients, we calculate an additional gradient for each model via codistillation. To apply codistillation, we first create a copy of the parameters of each model at the start of the round to act as a student. Let $\auxmodel^{\text{\tiny student}}$ and $\mainmodel^{\text{\tiny student}}$ denote the student models for $\auxmodel$ and $\mainmodel$, respectively. Using $\mainmodel^{\text{\tiny teacher}} = \mainmodel^t$ (respectively, $\auxmodel^{\text{\tiny teacher}} = \auxmodel^t$) as the teacher, we distill to $\auxmodel^{\text{\tiny student}}$ (respectively, $\mainmodel^{\text{\tiny student}}$). We then calculate the model difference between the model parameters at round $t$ and the student models at the end of the codistillation round, that is
\begin{equation}
\label{eq:dist-grad}
    \delta_i^t = \theta_i^t - \theta_i^{\text{\tiny student}},\,\, i \in \{L, H\}\,.
\end{equation}
We call the model difference induced by the distillation round the \emph{distillation gradient}. The last step involves merging the two gradients for each model: the aggregated \fedavg\  gradient and the distillation gradient. We can achieve this by a weighted combination of the two gradients using an interpolating factor $\alpha \in [0, 1]$. However, before merging, we ensure that the two gradients have the same scale by rescaling the distillation gradient. The final \emph{merged gradient} for each model can be written as
\begin{equation}
\label{eq:merged}
    \Delta^t_i = \alpha\, g^t_i + (1 - \alpha)\, \delta_i^t\,\frac{\Vert g_i^t\Vert_2}{\Vert \delta_i^t\Vert_2}\, ,\, i \in \{L, H\}\, ,
\end{equation}
where $\Vert\cdot\Vert_2$ denotes $\mathrm{L}_2$-norm. We then use the merged gradient for each model to perform the update and move on to the next \fedavg\ round. Note that in this setting, $\alpha = 1$ recovers the vanilla \fedavg\ approach. The \merged\ approach is summarized in Algorithm~\ref{alg:merged} and illustrated in Figure~\ref{fig:methods_illustrations}(b) in Appendix \ref{app:methods}.

%% file: 5_experiments.tex
\section{Experiments}\label{sec:experiments}
In this section, we provide experimental results. The details of the experimental setups are shown in Appendix \ref{app:setup}.

\subsection{\label{subsec:indom}Balanced Model Performance}

As we are interested in sharing information between the two server models, we first explore an experimental setup where the two models obtain similar performance and where knowledge distillation does not have to work against a large gap in performance between the teacher and the student. To accomplish this, we sample $\sim$700 clients from the StackOverflow dataset and $\sim$50 clients from the CIFAR-100 dataset as the high-capacity pools in their respective experiments. The data splits used are shown in Appendix \ref{app:setup}. %Table \ref{tab:data_equal}.
Acquiring in-domain unlabeled corpora for the distillation dataset may be even infeasible, but for many tasks, large out-of-domain corpora are often available. To evaluate how sensitive the distillation methods may be to distillation data domain shift, we run experiments with both in- and out-of-domain data. For image classification and language modeling, we use CIFAR-10 and C4-MOD, respectively, as the out-of-domain datasets. Results are shown in the middle column of Table \ref{tab:imbalance}. When the small and large models have similar performance \inter\ and \merged\ can improve performance over baseline \fedavg\ for both small and large models. The gains are greater when in-domain distillation data is available, but it is still possible to obtain gains using out-of-domain distillation data. The \merged\ technique yields gains in both tasks for both models. %Practically, it is often difficult to get in-domain labeled data in a federated learning setting, so we use in-domain unlabeled data for distillation in these initial experiments, in addition to experiments which make use of unlabeled out-of-domain data. Results are shown in Table \ref{tab:in_and_out}.

\begin{table*}[ht!]
\caption{\label{tab:imbalance}Accuracy on the test set, showing performance versus relative strength of the high-capacity model for both in- and out-of-domain distillation data. $|\mainpool|$ represents the size of the high-capacity pool in number of examples. %The middle column repeats the results of Table \ref{tab:in_and_out}.%
Best-performing models in each setup are in bold. The central column shows roughly balanced performance between the small and large models, as described in Section \ref{subsec:indom}.}
% \vskip 0.15in
%\vspace{-0.1cm}
\begin{center}
\begin{small}
\begin{sc}
\resizebox{1.5\columnwidth}{!}{%
\begin{tabular}{cc|cc|cc|cc}
\toprule
 & & \multicolumn{6}{c}{CIFAR} \\
 & in-domain & \multicolumn{2}{c}{ $|\mainpool|$=3100} & \multicolumn{2}{c}{ $|\mainpool|$=4700} & \multicolumn{2}{c}{ $|\mainpool|$=6300} \\
 & & small & large & small & large & small & large \\
\midrule
\fedavg & & 31.84 & 24.76 &  31.84 & 30.57 & 31.84  & 35.10  \\
\midrule
\inter & \checkmark &  \textbf{33.95} & 31.09 & \textbf{34.36} & \textbf{35.08} & \textbf{35.68} & \textbf{37.20}  \\
\merged & \checkmark &  32.86 & \textbf{31.98} & 33.86 & 33.15 & 33.78 & 33.96  \\
\midrule
\inter & x &  33.75 & 25.03 & 34.02 & 30.53 & 33.37 & 34.27 \\
\merged & x & 33.02 & 25.83 & 33.45 & 31.02 & 33.03 & 33.47 \\
\bottomrule
 & & \multicolumn{6}{c}{StackOverflow} \\
 & in-domain & \multicolumn{2}{c}{ $|\mainpool|$=143k} & \multicolumn{2}{c}{ $|\mainpool|$=296k} & \multicolumn{2}{c}{ $|\mainpool|$=612k} \\
 & & small & large & small & large & small & large \\
\midrule
\fedavg & x & 26.29 & 24.87 & 26.29 & 26.23  & 26.29 & 27.51  \\
\midrule
\inter & \checkmark & 25.70 & 26.26 & 25.95 & 26.68 & 25.73 & 27.06 \\
\merged & \checkmark & 26.42 & \textbf{26.63} & 26.50 & \textbf{26.72} & \textbf{26.66} & 27.55 \\
\midrule
\inter & x & 25.72 & 25.79 & 25.57 & 26.50  & 25.77 & 27.18  \\
\merged & x & \textbf{26.55} & 25.52 & \textbf{26.51} & 26.43 & 26.59 & \textbf{27.67}  \\
\bottomrule

\end{tabular}
}
\end{sc}
\end{small}
\end{center}
%\vspace{-0.4cm}
\end{table*}

\subsection{Imbalanced Model Performance}

% In the previous experiment, the performance of the large and small models is roughly comparable. However, in practice the large model may either outperform or underperform the small model. 
We now evaluate the extent to which the distillation methods are effective when the two models have imbalanced baseline performance. We achieve this setting by either increasing or decreasing the number of clients in the high-capacity pool. Results are shown in Table \ref{tab:imbalance}.
The most striking gains are seen when the small model is used as the teacher for the under-performing large model (Columns corresponding to $|\mainpool|$=3100 or 4700). In this case, the large model is limited by the size of its participating clients pool $\mainpool$; this allows us to leverage the small model trained on the low-capacity pool to make better use of the larger model's unutilized capacity. By sharing information bi-directionally via the distillation methods, we are able to improve the small model as well, via both methods for CIFAR and via \merged\ for StackOverflow. %While the technique is best motivated in the setting where the large model is under-performing, \merged\ makes gains over the \fedavg baseline for both models and tasks, and regardless of the domain of the distillation data everywhere except in the case of the large model for CIFAR when the large model is over-performing the small model. \textcolor{red}{this probably needs to be cut down}

% \fedavg & & 32.48 & 30.57 & 26.29 & 26.23  \\
% \midrule
% \inter & \checkmark & 34.36 & 35.08 & 25.95 & 26.68  \\
% \merged & \checkmark & 33.86 & 33.15 & 26.60 & 26.34 \\
% \midrule
% \inter & x & 34.02 & 30.53 & 25.57 & 26.50  \\
% \merged & x & 33.45 & 31.02 & 26.51 & 26.36 \\

% \begin{table*}
% \caption{\label{tab:in_and_out}Accuracy on the test set, using in and out of domain distillation data.}
% \vskip 0.15in

% \begin{center}
% \begin{small}
% \begin{sc}
% \begin{tabular}{cc|cc|cc|cc}
% \toprule
%  & & \multicolumn{6}{c}{CIFAR} \\
%  & in-domain & \multicolumn{2}{c}{n=31} & \multicolumn{2}{c}{n=47} & \multicolumn{2}{c}{n=63} \\
%  & & small & large & small & large & small & large \\
% \midrule
% \fedavg & & 31.84 & 24.76 &  31.84 & 30.57 & 31.84  & 35.10  \\
% \midrule
% \inter & \checkmark &  33.95 & 31.09 & 34.36 & 35.08 & 35.68 & 37.20  \\
% \merged & \checkmark &  32.86 & 31.98 & 33.86 & 33.15 & 32.43 & 33.42  \\
% \midrule
% \inter & x &  33.75 & 25.03 & 34.02 & 30.53 & 33.37 & 34.27 \\
% \merged & x & 33.02 & 25.83 & 33.45 & 31.02 & 33.03 & 33.47 \\
% \bottomrule

% \end{tabular}
% \end{sc}
% \end{small}
% \end{center}
% \end{table*}

\begin{table*}[h!]

\caption{\label{tab:domain_shift}Performance of \inter\ and \merged\ when the two \fedavg\ pools are domain-shifted. Test accuracy is reported, with Mixed representing the same evaluation set as used previously and Answers / Questions being split versions of that testset.}
% \vskip 0.15in
% \vspace{-0.1cm}
\begin{center}
\begin{small}
\begin{sc}
\resizebox{1.2\columnwidth}{!}{%
\begin{tabular}{c|cccccc}
\toprule

 & \multicolumn{6}{|c}{StackOverflow} \\
& \multicolumn{2}{c|}{answers} & \multicolumn{2}{c|}{questions} & \multicolumn{2}{c|}{mixed} \\
 & small & large & small & large & small & large  \\
\midrule
in-domain & \checkmark & X & X &\checkmark & $\sim$ & $\sim$ \\
\midrule
\fedavg & 26.00 & 24.79 & 22.62 & 30.34 & 24.51 & 27.25 \\
\inter & 25.56 & 24.82 & 22.51 & 30.23 & 24.21 & 27.23 \\
\merged & 25.50 & 25.29 & 27.65 & 30.75 & 26.45 & 27.70 \\
\bottomrule

\end{tabular}
}

\end{sc}
\end{small}
\end{center}
\vspace{-0.2cm}
\end{table*}

\subsection{Domain-shifted Client Pools}
% The previous experiments have modulated the size of the high-capacity pool $\mainpool$ without explicitly introducing domain-shift between the low-capacity and high-capacity pools. In practice, selecting subsets of clients from a federated pool can introduce biases in the models, in addition to impacting overall performance. 
We next introduce a domain-shift between the two \fedavg\ pools by splitting the StackOverflow examples such that the high-capacity pool $\mainpool$ only contains questions and the low-capacity pool $\auxpool$ only contains answers. Thus the pools are entirely distinct, as opposed to previous experiments when the high-capacity pool covered a subset of the clients in the low-capacity pool. As before, we use cross-domain distillation data, which contains both questions ($\sim45\%$) and answers ($\sim55\%$) to explore to what extent knowledge codistillation can help to address each model's weaknesses in their respective out-of-domain categories. We use the same test set as before and additionally evaluate on partitions of the test set, which contain only questions and only answers. The sizes of the pool are reported in Table \ref{tab:qa_data} in Appendix \ref{app:setup}. Results are reported in Table \ref{tab:domain_shift}.

% \merged\ works well to achieve domain transfer between the two models, yielding a $+5$ accuracy-point boost to the small model on the out-of-domain questions eval set. Likewise, the large model improves by $+0.5$ accuracy points on its out-of-domain answers eval. Interestingly the large model also improves on the in-domain questions, perhaps by incorporating the distilled knowledge from the small model to create a better overall language model. General performance improves for both models as well. \inter\ performed poorly compared to \fedavg\ in the domain-shifted experiments; plotting the performance of the methods over the course of the rounds in Figure \ref{fig:period_v_merged} makes clear that \inter\ has difficulty sharing information between the domain-shifted models because the student models never see information from both domains at once. Though \inter\ improves the out-of-domain performance through each codistillation round, seen in the upward spikes of the small model on Answers and the large model on Questions, these gains come at the cost of losing performance on the in-domain data. Subsequent \fedavg\ rounds re-establish the balance between the two domains without yielding additional gains beyond vanilla \fedavg. \merged\ does not have this issue because the model is shown information from both domains at all times, allowing an increase in general performance for both the small and large models.

\merged\ achieves domain transfer between the two models resulting in a $+5\%$ accuracy boost for the small model on out-of-domain questions and $+0.5\%$ accuracy boost for the large model on out-of-domain answers. The large model also improves on in-domain questions. Overall performance improves for both models, and \merged\ achieves faster convergence on the general performance (see Figure \ref{fig:period_v_merged} in Appendix \ref{app:experiments}). We analyze the poor performance of \inter\ in the domain-shifted setting in Appendix \ref{app:experiments}.

%% file: 6_discussion.tex
\section{Discussion}\label{sec:discussion}
Our experiments on the federated CIFAR-100 and StackOverflow datasets show that the effectiveness of the \inter\ and \merged\ strategies for codistillation depends on the sizes of the model, the client pool, and the distillation dataset. Codistillation offers the most benefit when the larger model, trained on the smaller subset of high-capacity clients, cannot utilize its full capacity due to limited training data. This is evident in Table~\ref{tab:imbalance} where the gains are more substantial for both codistillation strategies when the size of the high-capacity clients' pool is smaller. In most cases, both models can benefit by exchanging information throughout training; the larger model utilizes the information in the larger pool of clients through the smaller model, and the smaller model can benefit from the higher capacity of the larger model.
Of the two codistillation strategies, \inter\ is easier to integrate into standard \fedavg\ setups and does not require significant changes on the server side. Additionally, \inter\ offers more flexibility in terms of computation by adjusting the codistillation period based on the available resources. However, \merged\ is the more effective strategy and consistently improves performance in most scenarios. A downside of \merged\ is that it requires running codistillation at every federated round. To reduce this overhead of \merged, we can adjust the combination factor $\alpha$ (Equation~\ref{eq:merged}) dynamically during training. By setting $\alpha=1$, we can recover \fedavg\ and skip the codistillation cost for specific rounds. We plan to explore such strategies for more efficient utilization of computing resources in the future. In our experiments, we investigated codistillation in settings where the clients were partitioned according to capacity. Alternative criteria for partitioning clients could be based on user characteristics, an aspect we plan to explore in the future.

% Things to discuss: We've shown in the preliminary calibrating baselines that performance scales drastically with model size. The decision as to where to set the lowest-common-denominator model size that achieves the greatest performance for the greatest number of clients is not an obvious one. 

%  In Table \ref{tab:imbalance}, the most striking gains are seen when the small model is used as the teacher for the under-performing large model. This scenario is one where the high-capacity pool is limited not by the size of the model but by the size of the federated pool participating in \fedavg. This is the ideal scenario for distillation methods as we can leverage the small model trained on the low-capacity pool to make better use of the larger model's un-utilized capacity. In this setting, by sharing information bi-directionally via the distillation methods, we are able to improve the small model as well, via both methods for CIFAR and via \merged\ for StackOverflow. While the technique is best motivated in the setting where the large model is under-performing, \merged\ makes gains over the \fedavg baseline for both models and tasks, and regardless of the domain of the distillation data everywhere except in the case of the large model for CIFAR when the large model is over-performing the small model. \textcolor{red}{this probably needs to be cut down}

% In the experiments on domain-shifted StackOverflow (Table~\ref{tab:domain_shift}), 

% \textcolor{red}{TODO: explicitly discuss how \inter is a cheap version of \merged, easier to implement but worse overall.}

%% file: 7_conclusion.tex
\section{Conclusion}\label{sec:conclusion}

We have introduced two new methods for federated learning built on top of \fedavg; \inter\ and \merged, which address the unused modeling capacity on high-quality hardware necessitated by the shared-model-architecture constraint of \fedavg. We have conducted experiments on image classification and language modeling tasks and have shown that these methods outperform \fedavg. Both methods allow us to train a computationally efficient model on all clients while still benefiting from the enhanced modeling power of a larger model trained on a subset of the entire client pool. These methods are applicable even when only out-of-domain data is available or when only a limited amount of in-domain data can be collected. We have demonstrated that a high-capacity teacher can help improve a low-capacity student, even when the teacher has lower performance than the student. In addition to providing performance gains, both \merged\ and \inter\ are useful for sharing domain-specific knowledge between models. Our proposed methods provide a means to more fully utilize the modeling capacity available within a federated population. We will release code for both methods on publication of this work.

\newpage

%% file: 8a_related_work.tex
\section{Related Work}\label{app:related_work}

%\textcolor{red}{TODO}
\citet{mora2022knowledge} present an overview of Knowledge Distillation (KD) techniques for Federated Learning (FL). In one set of approaches, client models transfer knowledge to a centralized served model~\citep{seo2022}, In the second set, called \emph{co-distillation}~\citep{anil2018}, an ensemble of clients learn without a central server model. \citet{jeong2018communication} propose a distillation strategy that significantly reduces communication costs by exchanging soft targets instead of model parameters between the server and clients. In this approach, the clients transmit their per-label averaged logits computed on their private data, and the server averages these logits and broadcasts them in the next round. The clients then regularize their local loss using the server's logits. Variants of this method are presented by \citet{oh2020distill, ahn2019wireless, ahn2020cooperative}. \citet{itahara2020distillation} demonstrate that this technique does not perform as well as \fedavg\ on non-IID data in terms of communication cost and accuracy. They present a variation where clients compute logits on unlabeled data, which are then averaged to compute a global logit after applying a temperature parameter that is chosen to minimize the entropy of the softmax. Their strategy leads to a lower communication cost and better accuracy than \fedavg. \citet{afonin2021towards} present a theoretical framework for KD in FL, showing that the performance is limited by data heterogeneity. \citet{ozkara2021quped} propose a KD approach for learning personalized FL models where the clients can differ in terms of architecture and weight precision. \vspace{-0.12cm}

\textcolor{black}{
FedDF \citep{lin2020} allows for different model topologies between clients, rather than requiring all clients to use the same architecture. This is done by maintaining multiple server model prototypes, one for each unique client topology. In each round of federated learning, the client models are updated using their local data and an ensemble is computed by averaging logits from all client models sampled in that round, regardless of the architecture. This step enables sharing of information across model topologies. Distillation is then performed using the ensemble as the teacher and the server model prototype as the student. Each client model is sent the updated server model prototype that corresponds to its topology. Both FedDF and our method relax the assumption of identical model topology in FedAVG and enable sharing of information between models with different topologies. In our method, this is made possible by dividing the set of clients into pools based on capacity and using co-distillation to share information between the server pools. While FedDF diverges from \fedavg\ in using client logits for constructing server models, we use \fedavg\ as a building block. This makes our method easier to implement when compared to FedDF, which requires considerable changes to the FedAVG implementation. Unlike FedDF, our approach additionally enables learning an explicit shared representation within each client pool, whereas in FedDF, this is only possible in an indirect manner by initializing the server model prototype with the average of the client models with that topology.}
\textcolor{black}{By learning explicit shared representations within each client pool, our method is robust to data-heterogeneous scenarios, as each model can become an expert in its pool's domain, and be a useful teacher to the out-of-domain model(s). See Appendix \ref{app:domain_shift} for experiments demonstrating strong performance when splitting the clients introduces domain-shift between client pools.}\vspace{-0.12cm}
\citet{zhu2022} and \citet{diao2020heterofl} present FL approaches which are robust to heterogeneity in client sizes. In both approaches, each client updates a fraction of the parameters in the server model depending on its capacity. In \citet{zhu2022}, the model is learnt in an incremental manner by adding one parameter column at a time. At download (upload) time, a client only downloads (uploads) the first $k$ columns. In \citet{diao2020heterofl}., the parameters of the server model in each layer are decomposed into nested subsets. A client only receives the subset of the model depending on its capacity. The approach of \citet{zhu2022} is also resilient to network unreliability because if there is a disruption in the network after downloading the first $k$ columns, the remaining columns can be copied from the last round. Both these approaches and ours allow the possibility for different clients to have different network sizes. The difference between their work and ours is that in their case, there is no explicit strategy for the small (or large) capacity models to share representation amongst themselves. In contrast, we employ distinct pools for the small/large capacity clients which enables learning of a better shared representation within each pool\textcolor{black}{, and makes the server models more robust to data heterogeneity between pools}.  Like their approaches, our approach also benefits from communication efficiency in that the model sizes in the low-capacity pool are smaller than that in the high-capacity pool. In their approaches, the smaller subsets of global parameters benefit more from aggregation relative to larger subsets, which are thus relatively undertrained. Our approach does not share this limitation because it updates all parameters of the server model in each client pool. \citet{zhu2022} have an additional assumption that the model can be learnt in a progressive manner by adding one column at a time, but we do not rely on such an assumption. \textcolor{black}{Additionally, any method which trains a single server model can be used in any client pool of our methods as a drop-in replacement for \fedavg. This allows our method to subsume any gains made by FL methods which produce single server models, while retaining the advantages of model heterogeneity and pool-specific shared representations.}
\vspace{-0.12cm}

\textcolor{black}{
Many other methods make improvements to \fedavg\ while still training a single server model: 
Split-mix FL \cite{hong2022efficient} extends HeteroFL by splitting the server model into base models based on width, and allowing all base models to be trained on all clients.
FLANC \cite{mei2022resource} uses low rank approximation to assemble local models on the fly from a shared basis; each local model modifies the shared basis and the local model transformation coefficients. FedFTG \citep{zhang2022fine} fine-tunes the server model to correct the model shift after aggregation so that the server model preserves the information in the local models to the maximum extent. Other methods which modify \fedavg\ but do not address the constraint of model-heterogeneity include FedProx \cite{li2020federated}, FedMiD \cite{yuan2021federated}, and FedDual \cite{chen2022feddual}. Since the above approaches train a single server model and can be thought of as improvements over FedAVG, \textcolor{black}{our methods yield complementary and orthogonal gains as} our co-distillation framework can be applied in conjunction by using any of these methods as the federated algorithm applied to a single pool.
}

%% file: 8b_app_methods.tex
\newpage
\section{Methods}\label{app:methods}

\begin{algorithm}[h]
  \caption{\inter: Periodic Codistillation}
  \label{alg:periodic}
\begin{algorithmic}[1]
  \INPUT number of \fedavg\ rounds $T$, total clients pool $\pool$ and the subset of high-performance clients $\mainpool \subset \pool$, codistillation period $p$, codistillation steps $s$
  \\\vspace{-0.25cm}\hrulefill
  \vspace{-0.05cm}
  \STATE Initialize parameters $\auxmodel^0$ and $\mainmodel^0$
  \FOR{$t=1$ {\bfseries to} $T$}
  \STATE Apply \fedavg\ on $\auxmodel^t$ and $\mainmodel^t$ by sampling clients from $\pool$ and $\mainpool$, respectively
  \IF{$t\, \%\, p == 0$}
  \STATE Create \textbf{teachers} $\auxmodel^{\text{\tiny teacher}} \gets \auxmodel^t$ and $\mainmodel^{\text{\tiny teacher}} \gets \mainmodel^t$
  \STATE Initialize \textbf{students} $\uauxmodel^{\text{\tiny student}} \gets \auxmodel^t, \,\umainmodel^{\text{\tiny student}} \gets \mainmodel^t$
  \STATE \textbf{In parallel}, distill to $\theta^{\text{\tiny student}}_{i}$ for $s$ steps using $\theta^{\text{\tiny teacher}}_{\{L, H\} - \{i\}}$ as the teacher, where $i \in \{L, H\}$
  \STATE Replace $\auxmodel^t \gets \uauxmodel^{\text{\tiny student}}, \,\mainmodel^t \gets \umainmodel^{\text{\tiny student}}$
  \ENDIF
  \ENDFOR
  \OUTPUT $\auxmodel^T$, $\mainmodel^T$
\end{algorithmic}
\vspace{-0.1cm}
\end{algorithm}
% MOVED TO APPENDIX
\begin{algorithm}[h]
  \caption{\merged: Merged Codistillation}
  \label{alg:merged}
\begin{algorithmic}[1]
  \INPUT number of \fedavg\ rounds $T$, total clients pool $\pool$ and the subset of high-performance clients $\mainpool \subset \pool$, codistillation steps $s$, merging factor $\alpha \in [0, 1]$
  \\\vspace{-0.25cm}\hrulefill
  \vspace{-0.05cm}
  \STATE Initialize parameters $\auxmodel^0$ and $\mainmodel^0$
  \FOR{$t=1$ {\bfseries to} $T$}
  \STATE Aggregate client gradients $g^t_L$ and $g^t_H$, respectively, for $\auxmodel^t$ and $\mainmodel^t$ by sampling clients from $\pool$ and $\mainpool$
  \STATE Create \textbf{teachers} $\auxmodel^{\text{\tiny teacher}} \gets \auxmodel^t$ and $\mainmodel^{\text{\tiny teacher}} \gets \mainmodel^t$
  \STATE Initialize \textbf{students} $\uauxmodel^{\text{\tiny student}} \gets \auxmodel^t, \,\umainmodel^{\text{\tiny student}} \gets \mainmodel^t$
  \STATE \textbf{In parallel}, distill to $\theta^{\text{\tiny student}}_{i}$ for $s$ steps using $\theta^{\text{\tiny teacher}}_{\{L, H\} - \{i\}}$ as the teacher, where $i \in \{L, H\}$
  \STATE Calculate \textbf{distillation gradients}\vspace{-0.3cm}\[\delta_i^t \gets \theta_i^t - \theta_i^{\text{\tiny student}},\,\, i \in \{L, H\}\]\vspace{-0.7cm}\\
  \STATE \textbf{Merge} the gradients\vspace{-0.5cm}\[\Delta^t_i \gets \alpha\, g^t_i + (1 - \alpha)\, \delta_i^t\,\frac{\Vert g_i^t\Vert_2}{\Vert \delta_i^t\Vert_2}\, ,\, i \in \{L, H\}\]\vspace{-0.4cm}\\
  \STATE Apply \fedavg\ on $\theta^t_i$ using $\Delta^t_i$ as the gradient to find $\theta^{t+1}_i$, where $i \in \{L, H\}$
  \ENDFOR
  \OUTPUT $\auxmodel^T$, $\mainmodel^T$
\end{algorithmic}
\vspace{-0.1cm}
\end{algorithm}
% MOVED TO APPENDIX

\begin{figure*}[b!]
\begin{center}
\vspace{-2cm}
    % \subfigure[]{\includegraphics[width=0.49\linewidth]{figs/fedavg.pdf}}
    % % \hspace{-0.5cm}
    % \subfigure[]{\includegraphics[width=0.49\linewidth]{figs/codist.pdf}}
    \hspace{0.3cm}
    \subfigure[]{\includegraphics[width=0.35\linewidth]{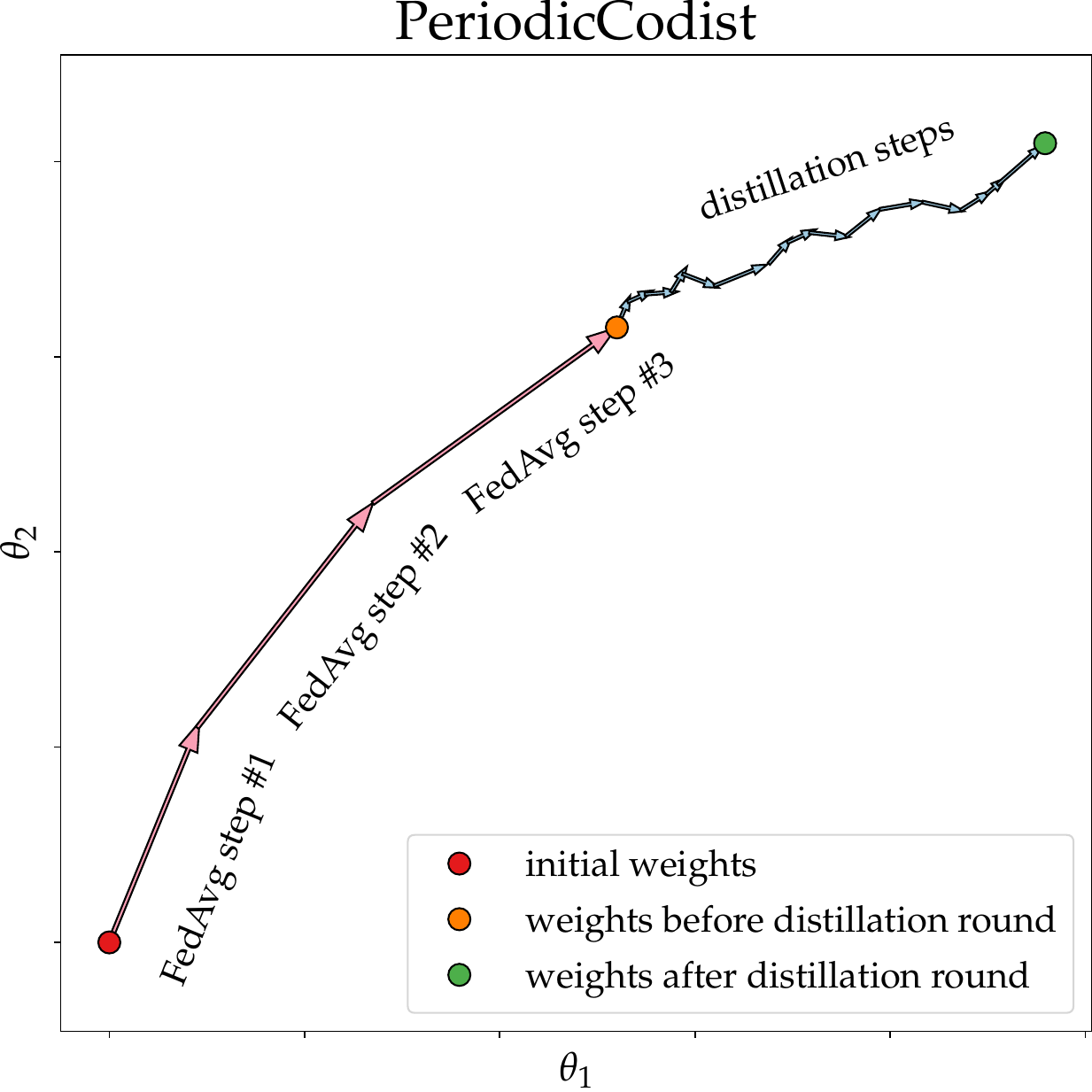}}
    % \hspace{0.1cm}
    \subfigure[]{\includegraphics[width=0.35\linewidth]{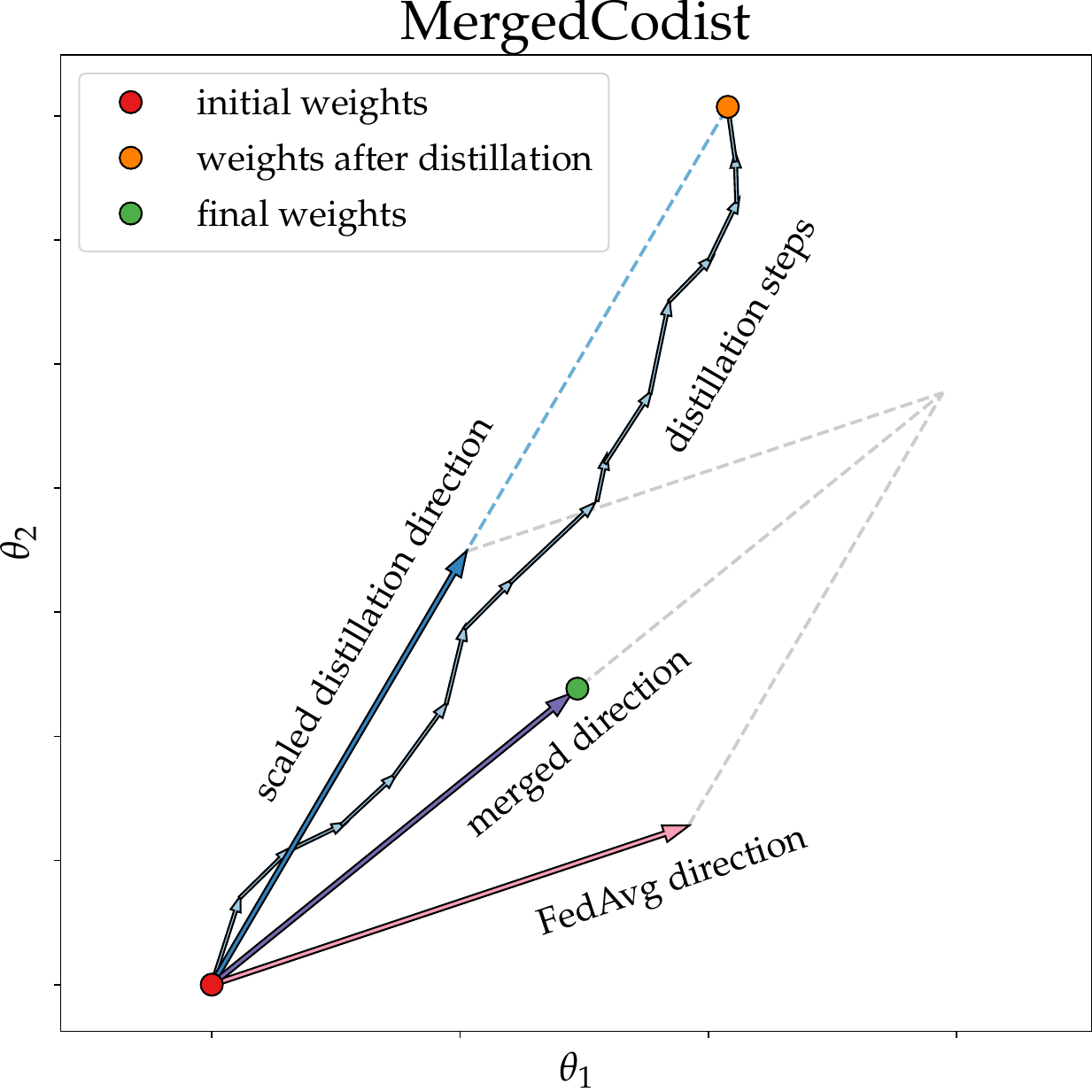}}
    \vspace{-0.3cm}
    \caption{
    %Schematics of the proposed methods: (a) Baseline \fedavg\ on the single pool of clients. The model capacity is limited by the lowest-common-denominator client's capacity in the pool. (b) Federated co-distillation with two client pools. The larger model trains on a smaller pool of high-capacity clients. The models exchange information via co-distillation on the server. 
    (a) An illustration of the \inter\ approach in the weight space. After $p=3$ \fedavg\ steps, we apply a round of co-distillation for $s=15$ steps by using the other model as a teacher. The \fedavg\ resumes after the co-distillation round. (b) An illustration of the \merged\ approach. In each round, we merge the scaled distillation direction (by calculating the distillation direction after $s=15$ steps) with the \fedavg\ direction before applying the update to the model.}
    \label{fig:methods_illustrations}
    \vspace{-0.3cm}
    \end{center}
\end{figure*}

%% file: 8c_experiment_setup.tex
\newpage
\section{Experiment Setup}\label{app:setup}

\begin{table*}[h]
\caption{Dataset sizes used for the experiments. The distillation datasets for federated CIFAR-100 and StackOverflow are excised from the original datasets' training split, as is the held-out set for CIFAR-100. StackOverflow has default held-out and test sets of $\sim$16M examples each. From each, we draw 5,000 and 100,000 examples for tuning and evaluation, respectively.}
\label{tab:data}
% \vskip 0.15in
\vspace{-0.1cm}
\begin{center}
\begin{small}
\begin{sc}
% \resizebox{1.0\columnwidth}{!}{%
\begin{tabular}{l|c|cc|c|r}
\toprule
\multirow{3}{0em}{Dataset}  & Distillation & \multicolumn{2}{ c }{Federated Pool} & \multicolumn{1}{| c }{Held Out}& \multicolumn{1}{| r }{Test} \\
& Examples & Clients & Examples & Examples & Examples \\
\midrule
CIFAR-100   & 13,500  & 315  & 31,500  & 5,000  & 10,000 \\
CIFAR-10    & 50,000   &   &  &  \\
\midrule
% StackOverflow    & & 342,477 & 135,818,730 & 38,758 & 16,491,230 & 204,088 & 16,586,035 \\
StackOverflow    & 122,250,952  & 34,248 & 13,567,778 & 100,000 & 100,000 \\
C4-mod    &  4,488,694 &   &  & &  \\
% \multirow{3}{0em}{Dataset}  & Distillation & \multicolumn{2}{ c }{Federated Pool} & \multicolumn{2}{| c }{Held Out}& \multicolumn{2}{| c }{Test} \\
% & Examples & Clients & Examples & Clients & Examples & Clients & Examples \\
% \midrule
% CIFAR-100   & 13,500  & 315  & 31,500 & 50  & 5,000 & 100 & 10,000 \\
% CIFAR-10    & 50,000   &   &  &  &  &  &  \\
% \midrule
% % StackOverflow    & & 342,477 & 135,818,730 & 38,758 & 16,491,230 & 204,088 & 16,586,035 \\
% StackOverflow    & 122,250,952  & 34,248 & 13,567,778 & 38,758 & 16,491,230 & 204,088 & 16,586,035 \\
% C4-mod    &  4,488,694 &   &  &  &  &  &  \\
\bottomrule
\end{tabular}
% }
\end{sc}
\end{small}
\end{center}
\vskip -0.15in
\end{table*}

To evaluate \inter\ and \merged, we run experiments on the tasks of image classification and language modeling, comparing against a baseline of \fedavg. For image classification experiments, we use the federated version of the CIFAR-100 dataset\footnote{\tiny \url{https://www.tensorflow.org/federated/api_docs/python/tff/simulation/datasets/cifar100}} \cite{krizhevsky2009learning}, supplemented by CIFAR-10 as out-of-domain distillation data. For language modeling, we use federated Stack Overflow (SO)\footnote{\tiny{\url{https://www.tensorflow.org/federated/api_docs/python/tff/simulation/datasets/stackoverflow}}}, which we supplement with out-of-domain distillation data drawn from the C4 \cite{raffel2020exploring} dataset and modified to match the casing and tokenization of the SO data; we refer to this as C4-MOD. All dataset sizes are shown in Table~\ref{tab:data}. For both CIFAR-100 and SO datasets, we hold out 10\% of the original train split as validation data.

All experiments are run in the FedJax framework~\cite{fedjax}, and all hyper-parameters are tuned on the held-out sets through the Vizier hyper-parameter tuning service\footnote{\tiny \url{ https://cloud.google.com/ai-platform/optimizer/docs/overview}} \cite{golovin2017google} with default settings, to a max of 1024 trials.\footnote{Details of hyper-parameter tuning are shown in Appendix~\ref{app:tuning}.} As all experiment runs have both `small' (low-capacity) and `large' (high-capacity) server models, we tune hyper-parameters to maximize the average accuracy of the two models on the held-out set.

\begin{table*}[h!]
\vspace{-0.2cm}
\caption{\label{tab:data_equal}Data splits for the initial set of experiments.}
% \vskip 0.15in
\begin{center}
\begin{small}
\begin{sc}
% \resizebox{1.0\columnwidth}{!}{%
\begin{tabular}{c|c|cc|cc}
\toprule

 & distillation data & \multicolumn{4}{c}{federated pools}\\ % & \multicolumn{2}{|c}{SO} \\
 &  & \multicolumn{2}{c|}{full pool ($\pool$)} & \multicolumn{2}{c}{high-capacity ($\mainpool$)}  \\%& small & large \\
 & examples & clients & examples & clients & examples \\%& small & large \\
\midrule
CIFAR & 13,500  & 315 & 31,500 & 47 & 4700 \\%& 26.29 & 26.82  \\
StackOverflow  & 122,250,952  & 34,248 & 13,567,778 & 684 & 296,597 \\%& 26.1 & 26.62  \\
\bottomrule
\end{tabular}
% }

\end{sc}
\end{small}
\end{center}
\vspace{-0.3cm}
\end{table*}

\subsection{Image Classification}
For the CIFAR-10 and CIFAR-100 datasets, we pre-process all examples with the default settings in FedJAX; all images are centered and normalized, and training images are randomly cropped and horizontally flipped. When using the data for distillation, we apply additional random flipping, which may be vertical or horizontal, and mixup~\cite{mixup} with $\beta=2$. 

For all image classification experiments, we train Convolutional Neural Network (CNNs)~\citep{goodfellow2016deep} with five layers; for the model trained on the high-capacity pool, hereon the `large' model, we train with approx 410k parameters. For the model trained on the low-capacity pool, hereon the `small' model, we train with approx 109k parameters. The details of the model architectures are given in Appendix~\ref{app:model_architectures}.

\subsection{Language Modeling}
For the StackOverflow dataset, we tokenize with a lower-cased 4k word-piece model~\citep{wordpieces}\ fit to the training split. To render the C4 data in the same format as the SO data, we pre-process the C4 dataset by lower-casing and word-tokenizing the first sentence from each C4 example paragraph.\footnote{Word and sentence tokenization via NLTK \cite{bird2009natural}} We additionally correct the remaining differences to the SO tokenization by replacing a small hand-coded list of incorrectly tokenized contractions.\footnote{This list is shown in Appendix~\ref{appendix:contractions}} As the full C4 dataset is substantially larger than what we require for distillation data, we draw from it a subset of 4.5M examples. We train LSTMs~\cite{lstm} with one layer for the `small' model and two layers for the `large' model. 
For the `large' model, we train with approx 4.3M parameters and the `small' with approx 2M parameters. The details of the model architectures are given in Appendix \ref{app:model_architectures}.

For the language modeling runs, rather than using the entire default held-out and test sets with $\sim$16M examples, we draw 100k examples from each. We tune hyperparameters on the held-out set and report accuracy on the test set. To simulate a practical federated learning setting that is constrained in both modeling capacity and data availability, in all experiments, we use only a subset of the federated training pool of the StackOverflow dataset, leaving the remainder of examples unused, or for use as in-domain distillation data.

\subsection{Federated Averaging Baseline}
For all experiments, we use a set of shared \fedavg\ settings. All experiments are run for 1500 federated rounds, sampling 20 clients per round. The clients are optimized with SGD with a constant learning rate. We set the client batch size to 20; all clients train locally for one epoch. We optimize the server models with FedAdam \cite{kingma2014adam,reddi2020adaptive}, using the default hyper-parameters,\footnote{$\beta_1=$\,\,0.9, $\beta_2=\,\,$0.999, $\epsilon=$\,\,1e-5} and with separate tuned learning rates for the large and small models.  We use a linearly decaying learning rate (with a zero final value) over the course of the federated rounds.

\subsection{Distillation}

For each teacher model, we tune a softmax temperature parameter $T > 0$
which may either sharpen (when $T < 1$) or smooth (when $T > 1$) the output probability distribution: $p(y_i) = \frac{\exp\left(\frac{z_i}{T}\right)}{\sum_{j} \exp\left(\frac{z_j}{T}\right)}$, where $y_i$ is a label and $z_i$ is the corresponding logit. We additionally tune a student regularization term which modifies the target distribution by interpolating between the temperature-scaled teacher distribution and the soft-labels of the initial parameters of the student. This corresponds to using both $\theta^{\text{\tiny teacher}}_{\{L, H\} - i}$ and $\theta^{\text{\tiny teacher}}_{i}$ as teachers, when distilling to $\theta^{\text{\tiny student}}_{i}, \, i \in \{L, H\}$. The former teacher is used to predict soft-labels for distillation, while the latter  acts as a regularizer, preventing the student model from drifting too far from the parameters at the start of the distillation round.

We use the Kullback–Leibler divergence~\citep{murphy2022probabilistic} between the student probabilities and the target probabilities as the distillation loss. For both \merged\ and \inter, we optimize the distillation updates with Adam using the default hyper-parameters. For each student optimizer, we tune a learning rate that decays linearly over all rounds.

Though either method can be trivially extended to incorporate labeled distillation data by tuning an additional interpolating value between the teacher soft-labels and the true one-hot labels, in all experiments, we use only unlabeled distillation data. Our goal is to avoid any gains by exposing the models to additional labeled data and, thus, demonstrate the improvements solely due to codistillation. Nonetheless, unlabeled data is easier to acquire in practice than labeled data, especially in our domain of interest.
 
% \subsubsection{\inter}
In all experiments with \inter, we set the codistillation period to $p=200$ rounds and run codistillation for $s=200$ steps each codistillation round. When running \inter, we reset the momentum variables of the \fedavg\ optimizers after each codistillation round, as the models have departed from their initial parameters, rendering the momentum stale. When running \merged, we take $s=32$ distillation steps to form each round's distillation gradient. %When combining the \fedavg\ and distillation gradients in \merged, we re-scale the norm of the merged gradient to the size of the original \fedavg\ gradient. 

\subsection{Domain-shifted Client Pools}

\begin{table}[h]
\vspace{-0.3cm}
\caption{\label{tab:qa_data}Dataset sizes for the StackOverflow domain-shift  experiments. Distillation data is $\sim$55\% answers and  $\sim$45\% questions, reflecting the split in the overall federated StackOverflow dataset.}
% \vskip 0.15in
\vspace{-0.1cm}
\begin{center}
\begin{small}
\begin{sc}
% \resizebox{0.9\columnwidth}{!}{%
\begin{tabular}{c|cc}
\toprule
 & clients & examples \\
 \toprule
%  distillation data & - & 4,126,929 \\
%  \midrule
 answers only ($\auxpool$) & 30,000 & 11,580,368 \\
 questions only ($\mainpool$) & 30,000 & 5,060,866 \\
 \bottomrule
%  & distillation data & \multicolumn{4}{c}{federated pools}\\ % & \multicolumn{2}{|c}{SO} \\
%  &  & \multicolumn{2}{c|}{Answers only ($\auxpool$)} & \multicolumn{2}{c}{Questions only ($\mainpool$)}  \\%& small & large \\
%  & examples & clients & examples & clients & examples \\%& small & large \\
% \midrule
% StackOverflow  & 122,250,952  & 34,248 & 13,567,778 & 684 & 296,597 \\%& 26.1 & 26.62  \\
\end{tabular}
% }

\end{sc}
\end{small}
\end{center}
\vspace{-0.4cm}
\end{table}

%% file: 8d_experiments.tex
\newpage
\section{Further Analysis}\label{app:experiments}

\subsection{Domain-shifted Client Pools}\label{app:domain_shift}
\begin{figure*}[h]
% \vspace{-0.3cm}
\begin{center}
\includegraphics[width=16cm]{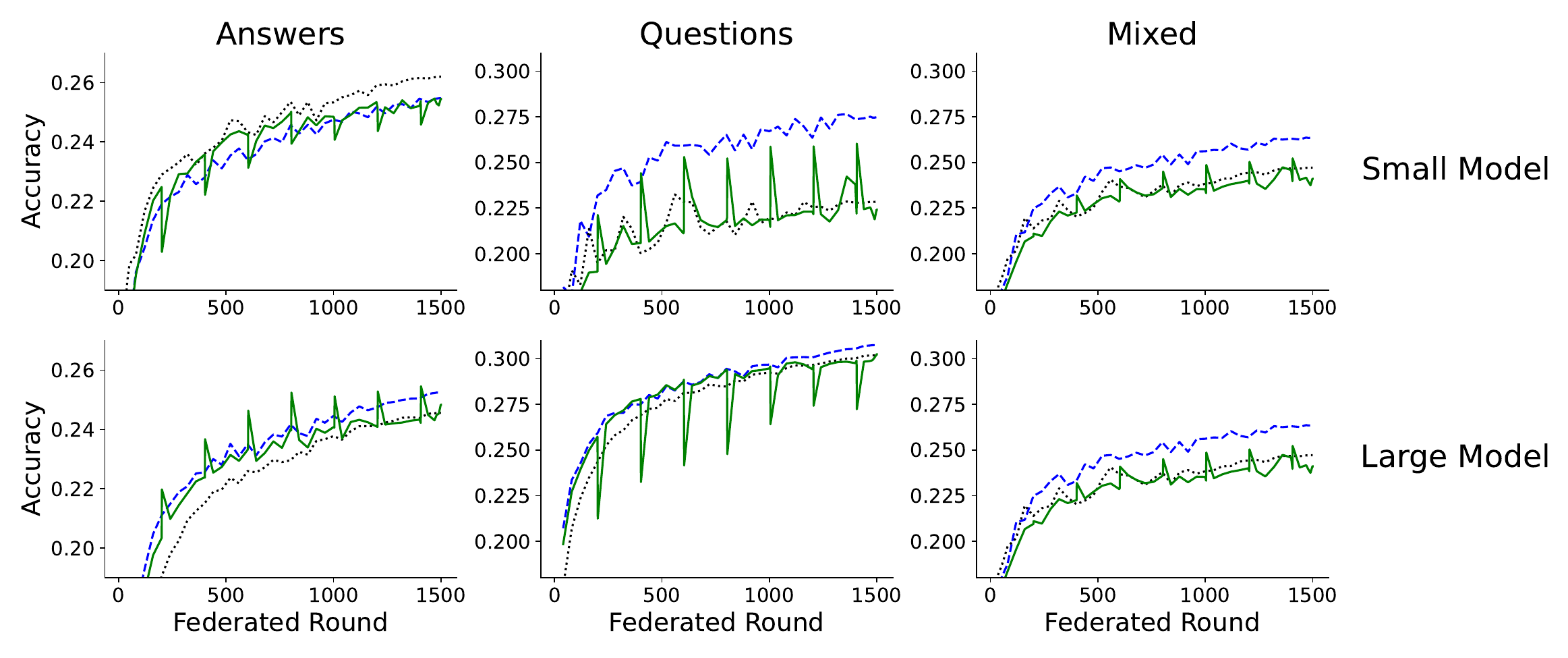}
\vspace{-0.3cm}
\caption{Performance of \merged\ in dashed blue (\textcolor{pythonblue}{\textbf{\texttt{--}}}) and \inter\ in solid green (\textcolor{pythongreen}{\textbf{---}}) vs. baseline of \fedavg\ in dotted black ($\bm{\cdot\,\cdot}$). Accuracy on the Answers-only, Questions-only, and Mixed test sets are shown. The small model \fedavg\ pool contains Answers only, and the large model pool contains Questions only. For \inter\, codistillation is run every $p=200$ rounds. Note the spikes in out-of-domain performance (large model on Answers and small model on Questions) following each codistillation round, and the corresponding drops in the in-domain performance.}\label{fig:period_v_merged}
\end{center}
\vspace{-0.2cm}
\end{figure*}

\inter\ performs poorly compared to \fedavg\ in domain-shifted experiments.
Figure \ref{fig:period_v_merged} shows that it struggles to share information between models because the student model never sees information from both domains at once. In each codistillation round, it shows upward spikes on out-of-domain accuracy and downward spikes on in-domain accuracy, which are smoothed out by subsequent \fedavg\ rounds resulting in no net gains over \fedavg. \merged\ avoids this issue by showing the model information from both domains at all times, leading to overall improvement for both models, as well as faster convergence on the general performance.

\subsection{Distillation Dataset Size}

\begin{figure}[h]
% \vspace{-0.2cm}
\begin{center}
\includegraphics[width=9cm]{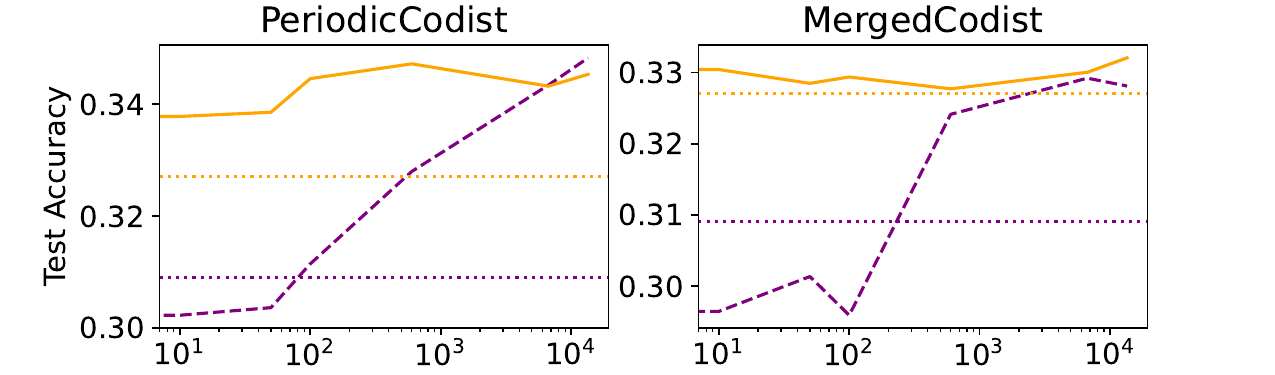}
\vspace{-0.2cm}
\caption{Performance of \inter\ and \merged\ on the CIFAR-100 dataset, with varying sizes of distillation data. The dashed purple line (\textcolor{violet}{\textbf{\texttt{--}}}) represents the large model, and the solid orange line (\textcolor{orange}{\textbf{---}}) the small model. The \fedavg\ baseline is shown as a dotted horizontal line for each model. For each distillation size, the averaged test accuracy of the 25 tuning runs with the best performance on the held-out set is shown. Though having more distillation data is better, good performance can be reached with just 600 examples for both methods.}
\label{fig:subdist}
\end{center}
\vspace{-0.2cm}
\end{figure}

In the experiments in Section \ref{subsec:indom}, we use as much in-domain data as is available. Realistically, acquiring even unlabeled in-domain data can be expensive, so available in-domain data may be limited in size. In Figure \ref{fig:subdist}, we evaluate how the performance of \inter\ and \merged\ varies as a function of the size of the distillation dataset, using the CIFAR-100 equally-performant setup from Section \ref{subsec:indom} as our baseline. Though \merged\ and \inter\ work best with more distillation data, it is possible to achieve good results on CIFAR-100 with only 600 unlabeled examples. This suggests that these methods may be practically useful even in settings where getting access to copious distillation data is expensive or infeasible.

%% file: 8z_appendix.tex
\newpage
\section{Model Architectures}\label{app:model_architectures}

For Image Classification, we train 5-layer CNNs with 3 convolutional layers and 2 dense layers. The small model has filter sizes $[16, 32, 32]$ and dense layer sizes $[64, 128]$, with a total of 109,348 parameters. The large model has filter sizes $[32, 64, 64]$ and dense layer sizes $[128, 256]$ with a total of 410,084 parameters. 

For Language Modeling, we train LSTMs with either one or two LSTM layers, following the modeling architecture described by \cite{reddi2020adaptive}. For the small model, we use an embedding size of 96 and LSTM hidden size of 70, for a total of 1,984,348 parameters. For the large model, we use an embedding size of 192 with a hidden size of 140, for a total of 4,278,644 parameters. 

All models are built in the haiku framework.\footnote{\url{https://dm-haiku.readthedocs.io/en/latest/}}

\section{Hyper-Parameter Tuning}\label{app:tuning}

For all runs, we tune hyper-parameters to maximize the accuracy on held-out data. We use the Vizier hyper-parameter tuning service \cite{golovin2017google} with the default setting, with 64 concurrent trials to a max of 1024 trials. For \fedavg\ runs, we tune a single client learning rate, and server learning rates for the small and large models. Learning rates are all tuned on the range $(0.0001, 0.1)$, which is explored on a log scale. When running \inter\, we additionally tune learning rates for the large model and small model distillation optimizers. We also tune decay coefficients for those learning rates, tuned on $(0,1)$ on a log scale. We further tune temperature coefficients on $(0.1,10)$ and server regularization interpolating variables on $(0.001, 0.5)$, both also on log scales. For \merged\ runs, we tune all of the above, and additionally tune the interpolating variable $\alpha$ on a linear scale on the range $(0.05,0.95)$.

% \twocolumn
% \newpage
\section{Contractions for C4-MOD tokenization}\label{appendix:contractions}

After sentence-tokenizing, word-tokenizing, and lower-casing the C4 data, we manually corrected the tokenization of punctuation with the following list of replacements:

\begin{multicols}{3}
\begin{itemize}[topsep=0pt,itemsep=0pt,partopsep=0pt, parsep=0pt]
    \item \texttt{aren ' t} $\rightarrow$ \texttt{aren't}
    \item \texttt{can ' t} $\rightarrow$ \texttt{can't}
    \item \texttt{couldn ' t} $\rightarrow$ \texttt{couldn't}
    \item \texttt{didn ' t} $\rightarrow$ \texttt{didn't}
    \item \texttt{doesn ' t} $\rightarrow$ \texttt{doesn't}
    \item \texttt{don ' t} $\rightarrow$ \texttt{don't}
    \item \texttt{hadn ' t} $\rightarrow$ \texttt{hadn't}
    \item \texttt{hasn ' t} $\rightarrow$ \texttt{hasn't}
    \item \texttt{haven ' t} $\rightarrow$ \texttt{haven't}
\item \texttt{he ' d} $\rightarrow$ \texttt{he'd}
\item \texttt{he ' ll} $\rightarrow$ \texttt{he'll}
\item \texttt{he ' s} $\rightarrow$ \texttt{he's}
\item \texttt{i ' d} $\rightarrow$ \texttt{i'd}
\item \texttt{it ' s} $\rightarrow$ \texttt{it's}
\item \texttt{i ' ll} $\rightarrow$ \texttt{i'll}
\item \texttt{i ' m} $\rightarrow$ \texttt{i'm}
\item \texttt{i ' ve} $\rightarrow$ \texttt{i've}
\item \texttt{isn ' t} $\rightarrow$ \texttt{isn't}
\item \texttt{let ' s} $\rightarrow$ \texttt{let's}
\item \texttt{mightn ' t} $\rightarrow$ \texttt{mightn't}
\item \texttt{mustn ' t} $\rightarrow$ \texttt{mustn't}
\item \texttt{shan ' t} $\rightarrow$ \texttt{shan't}
\item \texttt{she ' d} $\rightarrow$ \texttt{she'd}
\item \texttt{she ' ll} $\rightarrow$ \texttt{she'll}
\item \texttt{she ' s} $\rightarrow$ \texttt{she's}
\item \texttt{shouldn ' t} $\rightarrow$ \texttt{shouldn't}
\item \texttt{that ' s} $\rightarrow$ \texttt{that's}
\item \texttt{there ' s} $\rightarrow$ \texttt{there's}
\item \texttt{they ' d} $\rightarrow$ \texttt{they'd}
\item \texttt{they ' ll} $\rightarrow$ \texttt{they'll}
\item \texttt{they ' re} $\rightarrow$ \texttt{they're}
\item \texttt{they ' ve} $\rightarrow$ \texttt{they've}
\item \texttt{we ' d} $\rightarrow$ \texttt{we'd}
\item \texttt{we ' re} $\rightarrow$ \texttt{we're}
\item \texttt{we ' ve} $\rightarrow$ \texttt{we've}
\item \texttt{weren ' t} $\rightarrow$ \texttt{weren't}
\item \texttt{what ' ll} $\rightarrow$ \texttt{what'll}
\item \texttt{what ' re} $\rightarrow$ \texttt{what're}
\item \texttt{what ' s} $\rightarrow$ \texttt{what's}
\item \texttt{what ' ve} $\rightarrow$ \texttt{what've}
\item \texttt{where ' s} $\rightarrow$ \texttt{where's}
\item \texttt{who ' d} $\rightarrow$ \texttt{who'd}
\item \texttt{who ' ll} $\rightarrow$ \texttt{who'll}
\item \texttt{who ' re} $\rightarrow$ \texttt{who're}
\item \texttt{who ' s} $\rightarrow$ \texttt{who's}
\item \texttt{who ' ve} $\rightarrow$ \texttt{who've}
\item \texttt{won ' t} $\rightarrow$ \texttt{won't}
\item \texttt{wouldn ' t} $\rightarrow$ \texttt{wouldn't}
\item \texttt{you ' d} $\rightarrow$ \texttt{you'd}
\item \texttt{you ' ll} $\rightarrow$ \texttt{you'll}
\item \texttt{you ' re} $\rightarrow$ \texttt{you're}
\item \texttt{you ' ve} $\rightarrow$ \texttt{you've}
\item \texttt{ ' s} $\rightarrow$ \texttt{'s}
\end{itemize}
\end{multicols}